\documentclass{article}

\usepackage{PRIMEarxiv}

\usepackage[utf8]{inputenc} 
\usepackage[T1]{fontenc}    
\usepackage{hyperref}       
\usepackage{url}            
\usepackage{booktabs}       
\usepackage{amsfonts}       
\usepackage{nicefrac}       
\usepackage{microtype}      
\usepackage{lipsum}
\usepackage{fancyhdr}       
\usepackage{graphicx}       
\graphicspath{{media/}}     

\title{A Comprehensive Overview of Fish-Eye Camera Distortion Correction Methods
}

\author{
  Jian Xu \\
  SUSTech \\
  Southern University of Science and Technology \\
  ShenZhen, GuangDong\\
  \texttt{hwllo772@gmail.com} \\
  \And
  De-Wei Han \\
  SUSTech \\
  Southern University of Science and Technology \\
  ShenZhen, GuangDong\\
  \texttt{musthan@126.com} \\
  \And
  Kang Li \\
  SUSTech \\
  Southern University of Science and Technology \\
  ShenZhen, GuangDong\\
  \texttt{likangqd@163.com} \\
  \And
  Jun-Jie Li \\
  SUSTech \\
  Southern University of Science and Technology \\
  ShenZhen, GuangDong\\
  \texttt{lijj3@mail.sustech.edu.cn} \\
  \And
  Zhao-Yuan Ma \\
  SUSTech \\
  Southern University of Science and Technology \\
  ShenZhen, GuangDong\\
  \texttt{mazy@sustech.edu.cn} \\
}

\begin{document}
\maketitle

\begin{abstract}
The fisheye camera, with its unique wide field of view and other characteristics, has found extensive applications in various fields\cite{10.1007/s11263-007-0075-7,10.1007/s11263-006-0023-y}. However, the fisheye camera suffers from significant distortion compared to pinhole cameras, resulting in distorted images of captured objects. Fish-eye camera distortion is a common issue in digital image processing, requiring effective correction techniques to enhance image quality. This review provides a comprehensive overview of various methods used for fish-eye camera distortion correction\cite{hughes_ReviewGeometricDistortion_2008}. The article explores the polynomial distortion model, which utilizes polynomial functions to model and correct radial distortions. Additionally, alternative approaches such as panorama mapping, grid mapping, direct methods, and deep learning-based methods are discussed. The review highlights the advantages, limitations, and recent advancements of each method, enabling readers to make informed decisions based on their specific needs.
\end{abstract}


\section{Introduction}
Fish-eye lenses have gained popularity in various fields, including photography\cite{tehrani_CorrectingPerceivedPerspective_2016}, computer vision\cite{posada_FloorSegmentationOmnidirectional_2010}, robotics\cite{markovic_MovingObjectDetection_2014}, and virtual reality\cite{xiong1997creating}, due to their wide field of view and unique visual effects. However, these lenses often introduce significant distortion to the captured images, which can distort the shapes of objects and degrade image quality. To overcome this challenge, fish-eye camera distortion correction methods have been developed to rectify the images and restore their original appearance.

The correction of fish-eye camera distortion is a crucial task in digital image processing. It involves the application of mathematical models and algorithms to compensate for the non-linear distortions introduced by fish-eye lenses. Correcting the distortion can improve the accuracy of measurements, facilitate accurate object recognition, and enhance overall image quality for various applications.

This review aims to provide a comprehensive overview of the different methods employed to correct fish-eye camera distortion. The review will cover both traditional and more recent approaches, discussing their underlying principles, advantages, limitations, and potential applications. By understanding the various methods available, researchers, professionals, and enthusiasts in the field can make informed decisions about the most suitable technique for their specific needs.

The following sections will delve into the polynomial distortion model, which is widely used for fish-eye camera distortion correction. Additionally, alternative methods such as panorama mapping, grid mapping, direct methods, and deep learning-based approaches will be explored. Each method will be examined in detail, highlighting their strengths and weaknesses and providing insights into their practical implementation.

Overall, this review aims to serve as a valuable resource for individuals interested in fish-eye camera distortion correction. By presenting a comprehensive overview of the available methods, this review aims to facilitate a deeper understanding of the techniques involved and foster further advancements in the field of digital image processing.

\section{Camera Projection Models}
The imaging process of a fisheye camera is commonly approximated as a unit sphere projection model. The imaging process of a fisheye camera can be decomposed into two steps: first, linearly projecting the 3D points in space onto a virtual unit sphere; and then projecting the points on the unit sphere onto the image plane, which is a nonlinear process.
In the context of fisheye cameras, four common projection models are widely used: Equidistant Projection Model\cite{kingslake_HistoryPhotographicLens_1989}, Equiangular Projection Model\cite{kingslake_HistoryPhotographicLens_1989}, Orthographic Projection Model\cite{miyamoto_FishEyeLens_1964} and Stereographic Projection Model\cite{stevenson_NonparametricCorrectionDistortion_1996}.
\subsection{Equidistant Projection Model}
The Equidistant Projection Model assumes that the rays of light passing through the lens and projecting onto the image sensor form equal angles with the optical axis.
In this projection model, the mapping between 3D points (X, Y, Z) in the camera coordinate system and 2D image coordinates (u, v) can be expressed as follows:
\begin{equation}
  \theta =\arctan (Y,X)
\end{equation}
\begin{equation}
  \varphi =\arctan (\sqrt{X^2+Y^2},Z )
\end{equation}
\begin{equation}
  r_d =f*\varphi
\end{equation}
\begin{equation}
  u =r_d*\cos (\theta)
\end{equation}
\begin{equation}
  v =r_d*\sin (\theta)
\end{equation}

Here, ($\theta$, $\varphi$) represents the spherical coordinates on the unit sphere, $r$ is the radial distance from the optical center, ($u$, $v$) represents the normalized image coordinates, and $f$ is the focal length of the fisheye lens.

\subsection{Equiangular Projection Model}
The Equiangular Projection Model is commonly used for capturing panoramic or 360-degree images with fisheye lenses. It involves mapping the 3D points on a unit sphere to 2D image coordinates using an equiangular grid.
In this projection model, the mapping between 3D points ($X$, $Y$, $Z$) on the unit sphere and 2D image coordinates ($u$, $v$) can be expressed as follows:

\begin{equation}
  \theta =\arctan (Y,X)
\end{equation}
\begin{equation}
  \varphi =\arctan (\sqrt{X^2+Y^2},Z )
\end{equation}
\begin{equation}
  u =\theta+\pi /(2\pi)
\end{equation}
\begin{equation}
  v =(\theta+\pi /2)/\pi
\end{equation}
Here, ($\theta$, $\varphi$) represents the spherical coordinates on the unit sphere, and ($u$, $v$) represents the normalized image coordinates, ranging from 0 to 1.

\subsection{Orthographic Projection Model}
The Orthographic Projection Model is a camera projection model that assumes the rays of light from the scene are parallel and perpendicular to the image plane. In this model, the 3D points are directly projected onto a 2D image without any perspective distortion.
The mapping between the 3D points ($X$, $Y$, $Z$) in the camera coordinate system and the 2D image coordinates ($u$, $v$) can be expressed as follows:

\begin{equation}
  u =X/scale_x + center_x
\end{equation}
\begin{equation}
  v =Y/scale_y + center_y
\end{equation}
Here, ($u$, $v$) represents the image coordinates, ($X$, $Y$) represents the 3D points in the camera coordinate system.

\subsection{Stereographic Projection Model}
The characteristic of the Stereographic Projection Model is that it preserves angles, which is a desirable property in mathematics known as conformality. Preserving angles means that the angles formed by any intersecting lines remain unchanged after the transformation, even though the lines themselves may become curved. Under a conformal transformation, a circle still remains a circle (where a straight line can be considered a circle with an infinite diameter). Therefore, to some extent, a conformal transformation also preserves the "shape" of objects. In the simulated scenario below, all boundary lines on the surface of the cylinder are transformed into circular arcs, and all angles formed by intersecting lines remain unchanged at 90$°$.
\begin{equation}
  r_d =2f*\tan \frac{\theta}{2}
\end{equation}

\section{Distortion Correction Methods}
Camera distortion is the alteration of an image's perspective caused by the camera's lens, sensor, or other factors. There are several types of distortion, including: Radial distortion, Tangential distortion, as well as Non-linear distortion.

The purpose of camera distortion correction is to transform the distorted image captured by the camera into an image that resembles the ideal image produced by a pinhole camera. This correction aims to improve the accuracy of the image, enhance its visual quality, and meet the specific requirements of various applications. 

The application of fisheye cameras in computer vision often requires advanced distortion correction methods to ensure accurate and reliable image analysis. Fisheye lenses introduce significant distortions that can impact the accuracy of measurements, object recognition, and scene understanding. In this section, we discuss various state-of-the-art methods for fisheye camera distortion correction, aiming to transform the distorted fisheye images into rectified images resembling those captured by ideal pinhole cameras. 

\subsection{Distortion Types}

\subsubsection{Radial Distortions}
\textbf{Barrel Distortion:} In barrel distortion, straight lines near the edges of the image appear to curve outward, giving the image a barrel-like or fisheye effect. This distortion occurs because light rays passing through the outer parts of the lens bend more than those passing through the center.

\textbf{Pincushion Distortion:} Pincushion distortion is the opposite of barrel distortion. In pincushion distortion, straight lines near the edges of the image appear to curve inward, resembling the shape of a pincushion. This occurs when light rays passing through the center of the lens bend more than those passing through the outer parts.
\subsubsection{Tangential (De-centering) Distortions}
Tangential distortion, also known as de-centering distortion, arises when the lens assembly is not perfectly aligned with the image plane. This misalignment causes the image to be distorted in a non-radial manner, resulting in effects such as rotation and skewing. Unlike radial distortion, tangential distortion is not purely along the radial axis but varies with the radius from the image center.

\subsection{Polynomial Distortion Model}
The polynomial distortion model is one of the most commonly used methods for fish-eye camera distortion correction. It relies on a mathematical model that describes the radial distortion present in fish-eye images. Typically, this model uses polynomial functions to approximate the distortions and correct them. The correction process involves converting the pixel coordinates of the image to normalized coordinates and applying the polynomial functions to rectify the distortions. This method is widely adopted due to its simplicity and effectiveness in addressing radial distortions in fish-eye images. While there are more models than what is described here, the industry has largely standardized on the following two distortion models. 

\subsubsection{Brown-Conrady}
Brown-Conrady distortion\cite{Brown71close-rangecamera} is a widely used lens distortion model in computer vision and photogrammetry. The model is primarily employed to correct the geometric distortions caused by camera lenses, which can occur due to various factors such as imperfections in lens manufacturing, non-linearities in the imaging process, and physical limitations of the lens design.

The key components of the Brown-Conrady model include two aspects.

\textbf{Radial distortion (pincushion and barrel distortion):} Light rays bend more at points farther from the center of the lens than those closer to the center. This is expressed mathematically as:
\begin{equation}
  \delta r_s = k_{1s}r^3+k_{2s}r^5+k_{3s}r^7+...+k_{ns}r^{n+2}
\end{equation}
where $r$ is the distance from the center of the image, and $k_{1s}$, $k_{2s}$, $k_{3s}$, etc., are the radial distortion coefficients.

\textbf{Tangential distortion:} The lens is not perfectly aligned with the image plane, meaning there is an angle error between the sensor assembly and the lens. This is expressed mathematically as:
\begin{equation}
  \delta x = 2p_1xy+p_2(r^2+2x^2)
\end{equation}
\begin{equation}
  \delta y = p_1(r^2+2y^2)+2p_2xy
\end{equation}
where $x$ and $y$ are the image coordinates, and $p_1$ and $p_2$ are the tangential distortion coefficients.
After merging the mathematical models of both types of distortion, they can be represented by the following formula:
\begin{equation}
x_{corrected} = x_{distorted}(1 + k_1 r^2 + k_2 r^4 + k_3 r^6+\dots) + (2p_1 x_{distorted} y_{distorted} + p_2 (r^2 + 2x_{distorted}^2))
\end{equation}
\begin{equation}
y_{corrected} = y_{distorted}(1 + k_1 r^2 + k_2 r^4 + k_3 r^6+\dots) + (p_1 (r^2 + 2y_{distorted}^2) + 2p_2 x_{distorted} y_{distorted})
\end{equation}
where $x_{distorted}$ and $y_{distorted}$ are the distorted image coordinates, and $x_{corrected}$ and $y_{corrected}$ are the corrected image coordinates. The parameters $k_1$, $k_2$, $k_3$, $p_1$, and $p_2$ are the distortion coefficients that need to be estimated during the calibration process.

\subsubsection{Kannala-Brandt}
The Kannala-Brandt distortion model\cite{kannala_GenericCameraCalibration_2004} is also a mathematical model used to describe radial lens distortion. It's an extension of the classical Brown-Conrady distortion model and provides a more accurate representation of distortion, particularly for wide-angle lenses and fisheye lenses.

Kannala-Brandt summarized four models for fisheye imaging, along with the pinhole imaging model, and compared the distortion curves of each model.

The perspective projection of a pinhole camera can be described by the following formula:
\begin{equation}
  x = f tan \theta
\end{equation}
The four fisheye projection models are:
\begin{itemize}
  \item Stereographic projection
\begin{equation}
  x = 2f \tan \frac{\theta}{2}
\end{equation}
  \item Equidistant projection
\begin{equation}
  x = f \theta
\end{equation}
  \item Equiangular projection
\begin{equation}
  x = 2f \sin \frac{\theta}{2} 
\end{equation}
  \item Orthographic projection
\begin{equation}
  x = f \sin \theta
\end{equation}
\end{itemize}
\begin{figure}[!h]
  \centering
  \includegraphics[scale=0.3]{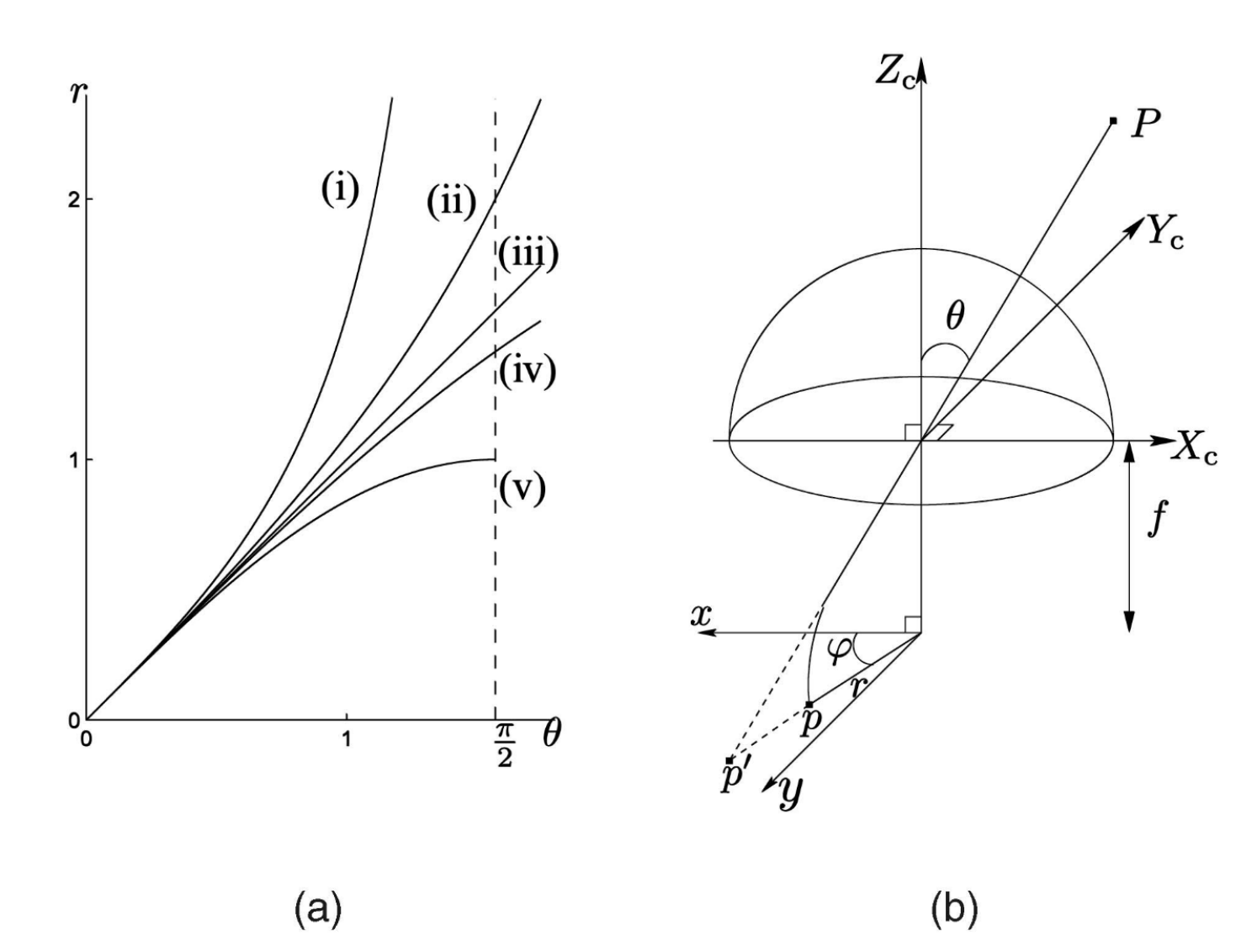}
  \caption{(a) Projections with f=1 ((i) to (v) are representing the five projection models mentioned above). (b) Fish-eye camera model.\cite{kannala_GenericCameraCalibration_2004} }
  \label{fig:1}
\end{figure}
\begin{equation}
  x = r(\theta) \cos \varphi 
\end{equation}
\begin{equation}
  y = r(\theta) \sin \varphi 
\end{equation}

The Kannala-Brandt model represents radial distortion using a polynomial function. Unlike the Brown-Conrady model, which typically uses a fixed number of polynomial terms (e.g., $k_1$, $k_2$, $k_3$), the Kannala-Brandt model employs a more flexible approach. It uses a radial polynomial with terms up to a certain order (typically denoted by $n$), allowing for a more accurate representation of complex distortion patterns.
To obtain a widely applicable, flexible model, they propose to use two distortion terms as follows: 
One distortion term acts in the radial direction.
\begin{equation}
  \delta r_s=(l_1\theta + l_2\theta ^3+l_3\theta ^5)(i_1\cos \phi + i_2\cos (2\phi)+i_4\sin (2\phi)+...)
\end{equation}
The other distortion term acts in the tangential direction.
\begin{equation}
  \delta t=(m_1\theta + m_2\theta ^3+m_3\theta ^5)(j_1\cos \phi + j_2\cos (2\phi)+j_4\sin (2\phi)+...)
\end{equation}
The complete camera model comprises 23 parameters, denoted as $p_{23}$ hereafter. Given the high flexibility of the asymmetric component of the model, there are situations where utilizing a simplified camera model may be preferable to prevent overfitting.

\subsection{Feature-based Methods}
In the context of fisheye camera distortion correction, feature-based methods play a vital role by leveraging the characteristics of the fisheye image to infer and rectify the camera's distortion parameters. This section presents an overview of several feature-based methods commonly used for fisheye camera distortion correction.
\subsubsection{Corner Detection and Rectification}
Corner detection and rectification methods involve the detection of corners in the fisheye image, such as employing Harris corner detection\cite{harris1988combined} or Shi-Tomasi corner detection\cite{shi1994good}. Subsequently, distortion rectification is performed by utilizing the relationships between the detected corners in terms of distances and angles to estimate the camera's distortion parameters. By employing these parameters, the entire image can be rectified to mitigate the fisheye distortion.

(Sixian Chan et al., 2016)\cite{7606985} present an enhanced automatic detection method for checkerboards, aiming to circumvent the constraints and user intervention typically associated with conventional methods. They evaluate a state-of-the-art corner detection technique, analyzing its advantages and limitations. To address the identified shortcomings, they develop an adaptive automatic corner detection algorithm.

(Zhang et al., 2020)\cite{zhang2020optimized} proposes an optimized calibration method tailored for ultra-wide dual-band cameras, leveraging a specifically designed thermal radiation checkerboard. The optimizations focus on subpixel location, center location, and mapping relationship refinement. Subpixel corner detection involves utilizing a variable quadrilateral template on edge gradients, with the minimum orthogonal value calculated to precisely locate the corners. Center location is achieved through Gaussian fitting, iteratively selecting the bottom point as the potential center. 

\subsubsection{Feature Point Matching and Rectification}
Feature point matching and rectification methods rely on the extraction and matching of feature points in the fisheye image. Popular techniques, such as SIFT feature points\cite{lowe2004distinctive} or SURF feature points\cite{bay2006surf}, are employed for feature point extraction. The detected feature points are then matched with their corresponding points in the rectified image, enabling the inference of the camera's distortion parameters and subsequent rectification.
\subsubsection{Line Detection and Rectification}
Line detection and rectification methods capitalize on the presence of straight lines in the fisheye image. By detecting lines using techniques like the Hough transform\cite{duda1972use}, the camera's distortion parameters can be inferred. The obtained parameters are subsequently utilized to rectify the entire image, ensuring that the straight lines preserve their linearity in the rectified image.

A groundbreaking study\cite{Devernay2001} introduced the concept that straight line segments in the physical world should retain their linearity even after being projected by a fisheye lens. Building upon this principle, Bukhari et al.\cite{Bukhari2013} proposed the utilization of an extended Hough transform of lines to correct radial distortions. 

(M. Zhang et. al., 2015)\cite{7299041} presents a novel algorithm for rectifying fisheye images in the undistorted perspective image plane by incorporating line constraints. They introduced a novel Multi-Label Energy Optimization (MLEO) method for merging short circular arcs with similar circular parameters and selecting long circular arcs to facilitate camera rectification.

(Zhu-Cun Xueet. al., 2020)\cite{xue2020fisheye} introduced a pioneering approach termed Line-aware Rectification Network (LaRecNet), aimed at tackling the challenge of fisheye distortion rectification. LaRecNet leverages the fundamental principle that straight lines in 3D space should retain their linearity in image planes.
\subsubsection{Optical Flow-based Methods}
Optical flow-based methods exploit the information provided by the optical flow in the fisheye image for distortion correction. By calculating the pixel displacements in the fisheye image, the camera's distortion parameters can be inferred and used for rectification. Optical flow algorithms, such as the Lucas-Kanade method\cite{lucas1981iterative} or deep learning-based optical flow estimation, can be employed to estimate the pixel displacements.

Existing methods often overlook the temporal correlation between frames, leading to temporal jitter in the corrected video. To tackle this issue, researchers propose a temporal weighting scheme that gradually reduces the weight of frames, resulting in a more coherent global optical flow (Shangrong Yang, et al., 2023)\cite{Yang_Lin_Liao_Zhao_2023}. By leveraging inter-frame optical flow, researchers can better perceive local spatial deformations in fisheye videos. Additionally, spatial deformation is derived using flows from both fisheye and distortion-free videos, enhancing the local accuracy of predictions. However, correcting each frame independently disrupts temporal correlation. To address this, a temporal deformation aggregator is introduced to reconstruct deformation correlation between frames and provide a reliable global feature\cite{Yang_Lin_Liao_Zhao_2023}.

\subsection{Direct Methods}
Direct methods for fish-eye camera distortion correction involve the detection and analysis of specific features or patterns in the image to estimate the distortion parameters. These methods typically rely on the relationships between the distorted image coordinates and the undistorted object coordinates. Techniques such as RANSAC (Random Sample Consensus)\cite{10.1145/358669.358692} can be used to robustly estimate the distortion parameters from the feature correspondences. Direct methods are advantageous in scenarios where calibration data or a priori knowledge about the distortion model is not available.
\subsubsection{Horizontal Expansion Method}
The Horizontal Expansion Method is a technique used for fisheye image rectification and distortion correction. It aims to transform a distorted fisheye image into a rectilinear image, which has straight lines and a more natural perspective. The method involves expanding the horizontal field of view of the fisheye image and mapping the distorted pixels to their corresponding locations in the rectilinear image.
\subsubsection{Latitude-Longitude Mapping Method}
The Latitude-Longitude Mapping Method was was first proposed by Kum et al. \cite{Latitude-Longitude2013}. This process resembles the mapping of the Earth's surface onto a plane. Initially, each image point is mapped onto a sphere within the latitude and longitude coordinate framework. Subsequently, the points along each meridian share the same abscissa value based on the latitude and longitude distribution, with greater distortion occurring as longitude values increase. Similarly, points along each latitude share the same ordinate value. However, the upper and lower regions of the fish-eye image often exhibit significant stretching and distortion due to excessive correction applied to the original image.
\subsubsection{Panorama Mapping Method}
The Panorama Mapping Method is primarily employed to achieve a complete 360° panoramic vista\cite{Xu_2019}. It entails expanding the field of view to capture the surrounding environment from all angles..

\subsection{Deep Learning-Based Methods}
With the recent advancements in deep learning, neural network-based approaches have emerged for fish-eye camera distortion correction. These methods involve training a neural network to learn the mapping function between distorted and undistorted images. A large dataset of paired images with known distortions is used for training the network. Once trained, the network can perform distortion correction on new input images. Deep learning-based methods offer the advantage of learning complex distortion patterns and can handle a wide range of distortions effectively. However, they require a substantial amount of training data and computational resources. This section provides an overview of deep learning methods commonly used for fisheye camera distortion correction.

\subsubsection{Converlutional Meural Metworks(CNNs)}
Convolutional Neural Networks (CNNs) have been extensively employed for fisheye camera distortion correction. These networks consist of multiple convolutional layers that extract hierarchical features from the input images. By training CNNs on a large dataset of distorted and undistorted fisheye image pairs, they can learn the underlying patterns and relationships to predict the distortion-free version of a given fisheye image.

(Rong et. al., 2016)\cite{rong2017radial} intends to employ CNNs (Convolutional Neural Networks), to achieve radial distortion correction. Inspired by the growing availability of image dataset with non-radial distortion (Rong et. al., 2016) propose a framework to address the issue by synthesizing images with radial distortion for CNNs.   

(Borkar et. al., 2019)\cite{borkar2019deepcorrect} evaluate the effect of image distortions like Gaussian blur and additive noise on the activations of pre-trained convolutional filters. (Borkar et. al., 2019) propose a metric to identify the most noise susceptible convolutional filters and rank them in order of the highest gain in classification accuracy upon correction.   Radial lens distortion often exists in images taken by commercial cameras, which does not satisfy the assumption of pinhole camera model. They generated images with a large number of images of high variation of radial distortion, which can be well exploited by deep CNN with a high learning capacity. 

(Shi et. al., 2018)\cite{shi2018radial} claim that a weight layer with inverted foveal models can be added to these existing CNNs methods for radial distortion correction.   

(Arsenali et. al., 2019)\cite{arsenali2019rotinvmtl} propose RotInvMTL: a multi-task network (MTL) to perform joint semantic segmentation, boundary prediction, and object detection directly on raw fisheye images.   An attempt to optimize a CNN-based detector for fisheye cameras was made, taking into consideration the barrel distortion, which complicates the object detection (Goodarzi et. al., 2019)\cite{goodarzi2019optimization}. The obtained result proves that fisheye augmentation can considerably advance a CNN-based detector’s performance on fisheye images in spite of the distortion. (Vasiljevic et. al., 2020)\cite{vasiljevic2020neural} show that self-supervision can be used to learn accurate depth and ego-motion estimation without prior knowledge of the camera model. 

(Vasiljevic et. al., 2020)\cite{vasiljevic2020neural} introduce Neural Ray Surfaces (NRS), convolutional networks that represent pixel-wise projection rays, approximating a wide range of cameras. (Ramachandran et. al., 2022)\cite{ramachandran2022woodscape} provide a detailed analysis on the competition which attracted the participation of 120 global teams and a total of 1492 submissions.   The fisheye image has a severe geometric distortion which may interfere with the stage of image registration and stitching. In the stage of fisheye image correction (Hao et. al., 2023)\cite{hao2023stronger} propose the Attention-based Nonlinear Activation Free Network (ANAFNet) to deblur fisheye images corrected by Zhang calibration method.  

(Yang et. al., 2021)\cite{yang2021progressively} propose Progressively Complementary Network which  focuses on developing an interpretable correction mechanism for fisheye distortion rectification networks and introduces a feature-level correction scheme. It incorporates a correction layer within skip connections and utilizes appearance flows from various layers to pre-correct image features. As a result, the decoder can more effectively reconstruct the final output using the remaining distortion-free information.

\subsubsection{Generative Adversarial Networks(GANs)}
Generative Adversarial Networks (GANs) have also been utilized for fisheye camera distortion correction. GANs consist of a generator network and a discriminator network, which are trained simultaneously in an adversarial manner. The generator network generates undistorted fisheye images, while the discriminator network aims to distinguish between the generated undistorted images and the real undistorted images. Through this adversarial training process, GANs can learn to generate high-quality undistorted fisheye images.

(Li et. al., 2012)\cite{li2012fish} present a novel embedded real-time fisheye image distortion correction algorithm with application in IP network camera. A fast and simple distortion correction method is introduced based on Midpoint Circle Algorithm (MCA) which aims to determine the pixel positions along a circle circumference based on incremental calculation of decision parameters.   

In (Aghayari et. al., 2017)\cite{aghayari2017geometric}, individual lens calibration is conducted, and the interior/relative orientation parameters (IOPs and ROPs) of the camera are established using a tailored calibration network applied to both central and side images captured by these lenses. Designed calibration network is considered as a free distortion grid and applied to the measured control points in the image space as correction terms by means of bilinear interpolation.   

(Nikonorov et. al., 2019)\cite{nikonorov2019deep} present a new end-to-end framework applying two convolutional neural networks (CNNs) to reconstruct images captured with multilevel diffractive lenses (MDLs). 

(Liao et. al., 2020)\cite{liao2019dr} present distortion rectification generative adversarial network (DR-GAN), a conditional generative adversarial network (GAN) for automatic radial DR. To the best of the knowledge, this is the first end-to-end trainable adversarial framework for radial distortion rectification.   
ß
(Gallego et. al., 2020)\cite{del2020blind} propose a method wherein three CNNs are trained in parallel, to predict a certain element pair in the 3*3 transformation matrix.  In order to improve the quality of low-light image (Zhang et. al., 2021)\cite{zhang2021novel} propose a Heterogenous low-light image enhancement method based on DenseNet generative adversarial network. Secondly, the feature map from low light image to normal light image is learned by using the generative adversarial network.   

(Thapa et. al., 2021)\cite{thapa2021learning} present the distortion-guided network (DG-Net) for restoring distortion-free underwater images. (Thapa et. al., 2021)\cite{thapa2021learning} then use a generative adversarial network guided by the distortion map to restore the sharp distortion-free image.   (Luo et. al., 2021)\cite{luo2022unsupervised} propose an unsupervised deep convolutional network that takes rectified stereo image pairs as input and outputs corresponding dense disparity maps. Second, the left and right images, which are reconstructed based on the input image pair and corresponding disparities as well as the vertical correction maps, are regarded as the outputs of the generative term of the generative adversarial network (GAN).

\subsubsection{Vision Transformer(ViT)}
Inspired by the significant success of transformer architectures in natural language processing (NLP), researchers have recently extended their application to computer vision (CV) tasks.
Vision Transformers (ViTs) have been proposed for fisheye camera distortion correction, leveraging the self-attention mechanism to capture long-range dependencies in the fisheye images. By training ViTs on a large dataset of distorted and undistorted fisheye images, they can learn the complex relationships between the distorted and undistorted pixels and perform distortion correction effectively.

Constrained by a fixed receptive field, existing methods have not fully exploited the global distribution and local symmetry of distortion. To address this, Fishformer\cite{yang2022fishformer} was introduced, processing fisheye images as sequences to enhance both global and local perception. The Transformer was fine-tuned based on the structural properties of fisheye images.

In fisheye images, distinct distortion patterns are regularly distributed across the image plane, regardless of the visual content, offering valuable cues for rectification. To leverage these cues effectively, SimFIR\cite{feng2023simfir} is introduced as a straightforward framework for fisheye image rectification through self-supervised representation learning. The technical process involves splitting a fisheye image into multiple patches and extracting their representations using a Vision Transformer (ViT).



\section{Experiments}
\subsection{Datasets}
Currently, there is no universally accepted fisheye distortion correction dataset. Therefore, the mainstream approach at present is to utilize generated datasets.

(Yang et. al., 2021)\cite{yang2022fishformer} select two completely different types of datasets:(1)Place2 dataset\cite{7968387} and (2)CelebA dataset\cite{liu2015deep}. They crop the image to be circular, and then leverage the four-parameters polynomial model to transform the normal image, generating fisheye images.

(Fan et. al., 2021)\cite{fan2021sir} evaluate the proposed SIR-Net on the synthetic datasets which contains 4 different datasets:ADE20k\cite{zhou2018semantic}, WireFrame\cite{huang2020learning}, COCO2017\cite{lin2015microsoft}, and Place365 dataset\cite{7968387}. They used three distortion models,i.e., FOV\cite{ViTAEv2}, Division Model\cite{990465}, EquiDistant distortion model\cite{Equidistant2010}, and synthesized the dataset for each distortion model separately.

(Nobuhiko Wakai et. al., 2021)\cite{wakai2021rethinking} used two large-scale datasets of outdoor panoramas called the StreetLearn dataset (Manhattan 2019 subset)\cite{mirowski2019streetlearn} and the SP360 dataset\cite{10.1145/3283289.3283348}. Figure 2 illustrates the random distribution of the training set generated by capturing image patches using camera models with the maximum incident angle $\eta _{max}$.
\begin{figure}[!h]
  \centering
  \includegraphics[scale=0.2]{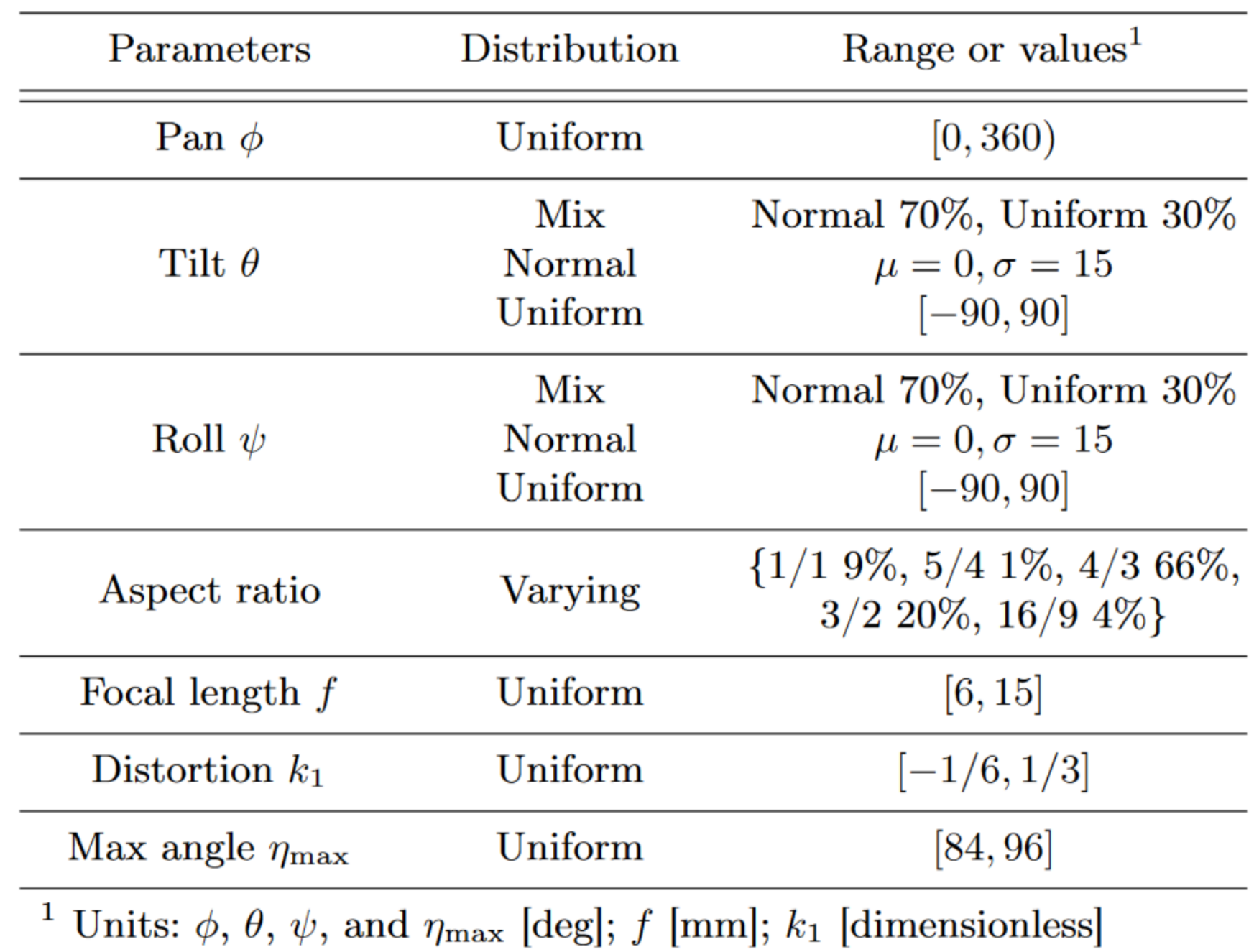}
  \caption{Distribution of the camera parameters for dataset.\cite{wakai2021rethinking} }
  \label{fig:2}
\end{figure}


\subsection{Evaluation Metrics}
Peak Signal to Noise Ratio (PSNR) and Structural Similarity (SSIM)\cite{1284395} stand as the two predominant evaluation metrics in image processing. PSNR provides an effective measure of the image's detailed quality, while SSIM offers an intuitive assessment of the image's structural integrity.


\section{Conclusion}
Fish-eye camera distortion correction is a critical task in digital image processing, aimed at rectifying the distortions introduced by fish-eye lenses and improving image quality. In this review, we provided a comprehensive overview of various methods used for fish-eye camera distortion correction.

We discussed the polynomial distortion model, which utilizes polynomial functions to model and correct radial distortions. This method is widely adopted due to its simplicity and effectiveness. Additionally, alternative approaches such as panorama mapping, grid mapping, direct methods, and deep learning-based methods were explored. Each method has its strengths and limitations, and their suitability depends on specific requirements and constraints.

Through this review, researchers, professionals, and enthusiasts in the field of digital image processing gained a deeper understanding of the available techniques for fish-eye camera distortion correction. The review highlighted the underlying principles, advantages, limitations, and potential applications of each method, enabling informed decision-making.

To evaluate the performance of distortion correction methods, various experiments can be conducted, including synthetic data evaluation, calibration image evaluation, comparative studies, real-time performance evaluation, and application-specific evaluations. These experiments provide insights into the accuracy, computational efficiency, and applicability of the methods in different scenarios.

In conclusion, fish-eye camera distortion correction methods play a crucial role in enhancing image quality and enabling accurate analysis in various fields. By understanding the different techniques and conducting appropriate experiments, researchers can select the most suitable method for their specific needs, contributing to advancements in the field of digital image processing.


\bibliographystyle{unsrt}  
\bibliography{Fisheye}

\end{document}